\documentclass[letterpaper, 10 pt, conference]{ieeeconf}  

\IEEEoverridecommandlockouts                              

\usepackage{mathptmx} 
\usepackage{times} 
\usepackage{bbding}
\usepackage{graphicx}
\usepackage{float}
\usepackage{diagbox}
\usepackage{arydshln} 
\usepackage{bbm}
\usepackage{wrapfig}
\usepackage{mathrsfs}
\usepackage{listings} 
\usepackage{physics}

\usepackage{graphics} 
\usepackage{float}
\usepackage{multirow}
\usepackage{amsmath}
\usepackage{bbm}
\usepackage{amssymb}
\usepackage{mathtools}
\usepackage[utf8]{inputenc} 
\usepackage[T1]{fontenc}    
\usepackage{hyperref}       
\usepackage{cleveref}
\usepackage{url}            
\usepackage{booktabs}       
\usepackage{amsfonts}       
\usepackage{nicefrac}       
\usepackage{microtype}      
\usepackage{xcolor}         
\usepackage{algorithm}
\usepackage{algorithmicx}
\usepackage{algorithm} 
\usepackage{algpseudocode}
\usepackage{fancyvrb}
\usepackage{fvextra}
\usepackage{csquotes}
\usepackage{threeparttable}

\newcommand{\zl}[1]{\textcolor{black}{#1}}
\newcommand{\zlw}[1]{\textcolor{black}{#1}}
\newcommand{\dingh}[1]{\textcolor{black}{#1}}

\title{\LARGE \bf
OPG-Policy: Occluded Push-Grasp Policy Learning with \\
Amodal Segmentation
}

\author{
Hao Ding, Yiming Zeng, Zhaoliang Wan, Hui Cheng
\thanks{This work was supported by the National Natural Science Foundation of China (U22A2095).}
\thanks{All authors are with the School of Computer Science and Engineering, Sun Yat-Sen University.
}
\thanks{
Corresponding to chengh9@mail.sysu.edu.cn}
}

\begin{document}
 

\newcommand{\sset}{\mathcal{S}}
\newcommand{\frees}{\mathcal{S}_f}
\newcommand{\aset}{\mathcal{A}}
\newcommand{\eps}{\epsilon}
\newcommand{\init}{\rho_0}
\newcommand{\E}{\mathbb{E}}
\newcommand{\R}{\mathbb{R}}
\newcommand{\hS}{\mathbb{S}}
\newcommand{\hP}{\mathbb{P}}
\newcommand{\M}{\mathcal{M}}
\newcommand{\bs}{\vb*{bs}}
\newcommand{\N}{\mathcal{N}}
\newcommand{\G}{\mathcal{G}}
\newcommand{\C}{\mathcal{C}}
\newcommand{\Env}{\mathcal{E}}
\newcommand{\x}{\vb*{x}}

\newcommand{\obj}{o}
\newcommand{\objs}{O}
\newcommand{\pose}{p}
\newcommand{\conds}{C}
\newcommand{\poses}{P}
\newcommand{\initpose}{\pose^0}
\newcommand{\initposes}{\poses^0}
\newcommand{\goalposes}{\poses^g}
\newcommand{\goalpose}{\pose^g}
\newcommand{\cond}{c}
\newcommand{\size}{s}
\newcommand{\cate}{y}
\newcommand{\mask}{m}
\newcommand{\obs}{I_{\objs}}
\newcommand{\func}{f}
\newcommand{\dist}{p_{\func}}
\newcommand{\data}{\textit{D}_{\func}}
\newcommand{\score}{\vb*{\Phi}_{\theta}}
\newcommand{\posespace}{\mathcal{P}}
\newcommand{\condspace}{\mathcal{C}}

\newcommand\mydata[2]{$#1_{\pm#2}$}
\newcommand{\indicator}{\mathbbm{1}}
\newcommand{\loss}{\mathcal{L}}
\newcommand{\trans}{\mathcal{T}}




\def\eg{\emph{e.g}.} \def\Eg{\emph{E.g}.}
\def\ie{\emph{i.e}.} \def\Ie{\emph{I.e}.}
\def\cf{\emph{c.f}.} \def\Cf{\emph{C.f}.}
\def\etc{\emph{etc}.} \def\vs{\emph{vs}.}
\def\wrt{w.r.t. } \def\dof{d.o.f. }
\def\etal{\emph{et al}. }

\maketitle

\begin{abstract}


Goal-oriented grasping in dense clutter, a fundamental challenge in robotics, demands an adaptive policy to handle occluded target objects and diverse configurations. Previous methods typically learn policies based on partially observable segments of the occluded target to generate motions. However, these policies often struggle to generate optimal motions due to uncertainties regarding the invisible portions of different occluded target objects across various scenes, resulting in low motion efficiency. To this end, we propose \textbf{OPG-Policy}, a novel framework that leverages amodal segmentation to predict occluded portions of the target and develop an adaptive push-grasp policy for cluttered scenarios where the target object is partially observed. Specifically, our approach trains a dedicated amodal segmentation module for diverse target objects to generate amodal masks. These masks and scene observations are mapped to the future rewards of grasp and push motion primitives via deep Q-learning to learn the motion critic. Afterward, the push and grasp motion candidates predicted by the critic, along with the relevant domain knowledge, are fed into the coordinator to generate the optimal motion implemented by the robot. Extensive experiments conducted in both simulated and real-world environments demonstrate the effectiveness of our approach in generating motion sequences for retrieving occluded targets, outperforming other baseline methods in success rate and motion efficiency.


\end{abstract}

\section{INTRODUCTION}



\begin{figure}[t]
\begin{center}
\includegraphics[width=\linewidth]{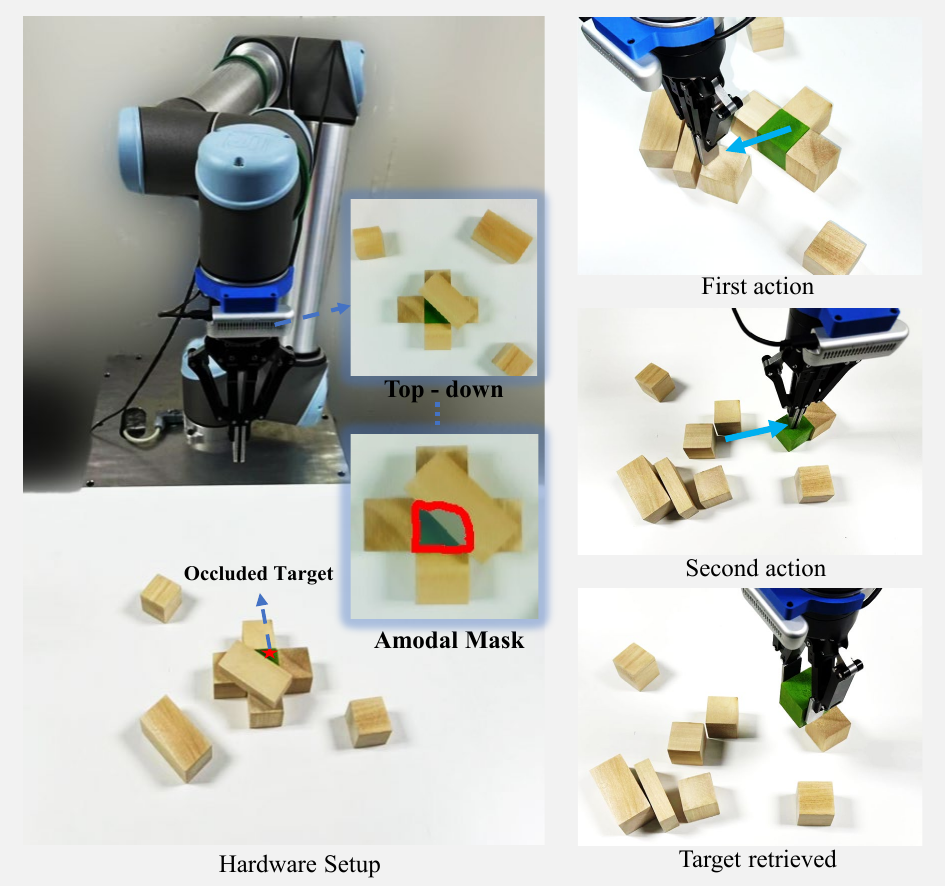}
\end{center}
\vspace{-10pt}
\caption{\textbf{Example configuration.} The target is the green cube occluded by a block on the top and surrounded by other blocks. We propose \textbf{OPG-Policy} to generate motion sequences with the assistance of amodal segmentation for occluded targets retrieval.}
\label{fig:teasor}
\vspace{-20pt}
\end{figure}

Goal-oriented grasping, a fundamental skill in robot manipulation, is extensively required for complex tasks in our daily lives such as table organization. As shown in Fig.~\ref{fig:teasor}, the target object is always occluded within a dense clutter, requiring the robot to present the ability to identify the occluded target and generate a motion sequence to efficiently clear the obstacles (\ie, push the target or surrounding objects) and grasp the target. Humans show remarkable abilities to accomplish this task with high motion efficiency and precision naturally, while it poses challenges for robotics since developing the manipulation policy with limited vision of target is difficult.

Previous works~\cite{xu2021efficient, yang2020deep, li2022learning, liu2022ge} employ different learning strategies to develop the manipulation policy between push and grasp. \cite{xu2021efficient} learns the policies via a DQN-based framework, focusing on visible portion of the target. Sampling-based method \cite{liu2022ge} trains an evaluator to select the optimal action from numerous sampling candidates. \cite{yang2020deep} designs a Bayesian-based policy for target exploration, then performs the coordination subtask based on the visible mask. Despite their success in handling cluttered scenarios, these methods suffer from low motion efficiency due to their inability to access information about the invisible portion of the target: there is no guarantee that the policy training will converge to the optimum with incomplete observations of the target object states.

To alleviate this issue and enhance the performance in cluttered scenarios, it is worth noting the potential benefits of incorporating the amodal segmentation module \cite{li2016amodal}, actively studied in Computer Vision, into goal-oriented grasping tasks. This approach enables the simultaneous segmentation of both visible and occluded parts of the target, offering a comprehensive understanding of its presence in the scene. 
Amodal segmentation has already shown remarkable capabilities in generic grasping \cite{wada2019joint,wada2018instance,inagaki2019detecting,qin2020s4g}, and exploring its potential in the context of goal-oriented grasping is promising. 

In this paper, we propose to introduce prior knowledge of target object shapes (\ie, amodal masks \cite{back2022unseen}) into a novel push-grasp policy training framework to address the issue of motion inefficiency in dense clutter. Firstly, a dedicated amodal segmentation module is trained for a group of target objects to generate amodal representations of the scenes. We can then design a reward to learn a motion critic via deep Q networks with inputs from the amodal representation, predicting push and grasp Q maps. By integrating the motions with the highest Q value and domain knowledge, we train a coordinator to determine the optimal action type based on current observation. In this manner, we develop a specific framework combining the strengths of amodal segmentation and reinforcement learning.

We conduct experiments in multiple dense clutter scenarios to demonstrate the effectiveness of our approach in improving motion efficiency and deploying them in real-world environments. Extensive results and analysis showcase that our method outperforms the baseline in total push/grasp attempts and success rate for varying configurations. Ablation studies further indicate that both amodal segmentation and the dedicated coordinator play indispensable roles.

The main contributions of this work are as follows:
\begin{itemize}
    \item We propose the Occluded Push-Grasp Policy, the first general learning framework for occluded goal-oriented grasping within a unified representation by treating the amodal segmentation masks as complementary inputs.
    \item We introduce a staged adaptive training method by designing a dynamically adaptive reward, which assists the push Q net to match the changes of the grasp Q net during the training process. We also design a coordinator based on amodal segmentation to generate the optimal action type based on the current workspace state.
    \item We conduct comprehensive experiments in simulation and real world without extra fine-tuning, demonstrating the effectiveness of our method in addressing dense clutter scenarios with high efficiency.
\end{itemize}

\section{RELATED WORK}
\subsection{\zl{Generic} Grasping}


\zlw{Generic grasping represents a fundamental robotic capability that has been extensively explored for decades. Initial studies \cite{kitaev2015physics,sahbani2012overview,bohg2013data} predominantly applied analytical methods aimed at identifying stable force closure grasp positions, relying on precise 3D models of objects to determine secure contact points between the objects and the grippers—a process often fraught with challenges. In contrast, recent advancements \cite{pmlr-v205-cai23a,newbury2023deep, xu2021adagrasp, DensePhysNet} have convincingly validated the potential of data-driven strategies, leveraging deep neural networks to bridge RGB or RGB-D imagery with grasp poses. These approaches, while effective in scenarios with sparsely located objects, frequently encounter limitations in densely cluttered environments, struggling to pinpoint viable grasping positions for specific targets.}

\begin{figure*}[t]
\begin{center}
\includegraphics[width=\linewidth]{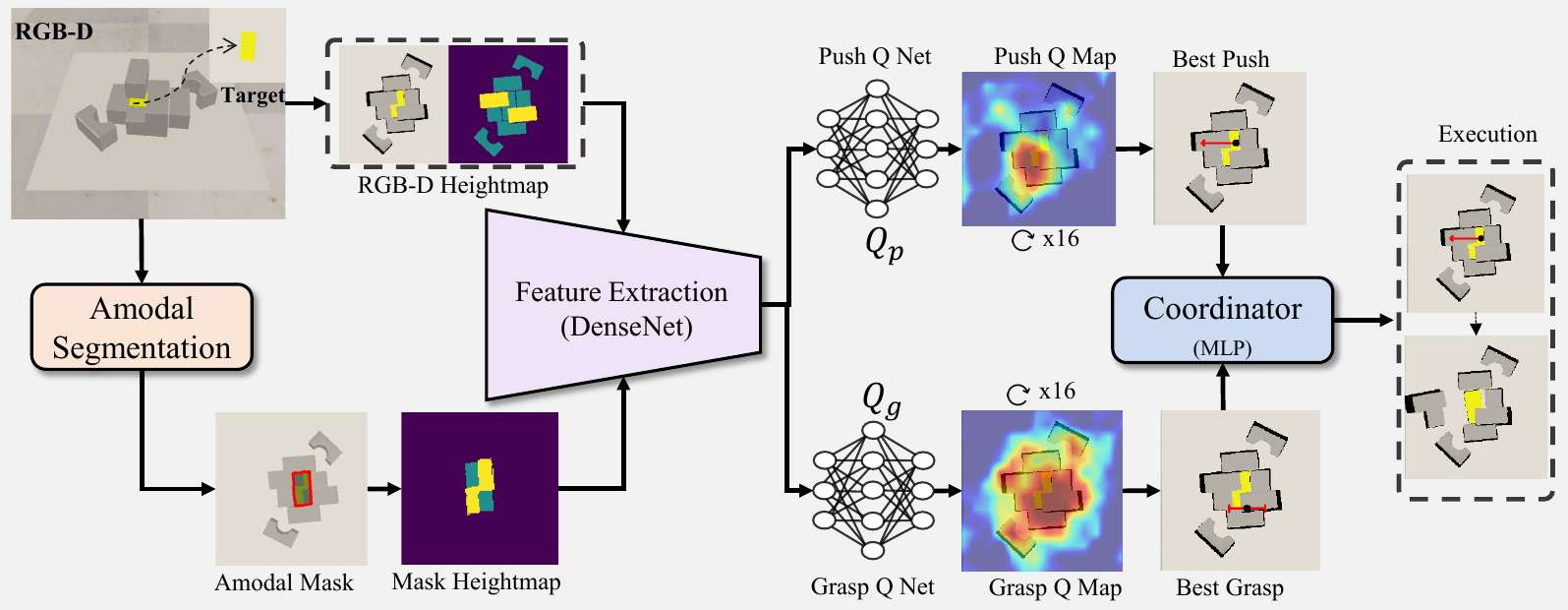}
\end{center}
\vspace{-10pt}
\caption{
\textbf{Occluded Push-Grasp Policy (OPG-Policy) Pipeline.} We fix a camera to capture RGB-D images of the workspace, which are then fed into the Amodal Segmentation Module to obtain the amodal mask of the severely occluded target. Next, RGB-D and amodal mask images are orthogonally projected towards gravity to get color, depth, and mask heightmaps. DenseNet processes these heightmaps to extract features for training the deep Q-networks, which predict push and grasp Q maps. These predictions and relative domain knowledge (which is not depicted in the figure) are then input into the coordinator to determine the optimal action from the best push and grasp.
}
\label{fig:pipeline}
\vspace{-15pt}
\end{figure*}

\subsection{\zl{Goal-oriented} Grasping}
\zlw{Goal-oriented grasping, aligning more closely with the complexities of real-world scenarios, necessitates the integration of non-prehensile actions, like pushing, to maneuver surrounding objects and create viable space for grasping. Studies by \cite{yang2020deep, xu2021efficient, kiatos2019robust, huang2022interleaving, yu2023iosg, zeng2024lvdiffusor} delve into the goal-oriented grasping. Yang et al. \cite{yang2020deep} introduce a Bayesian exploration approach to unveil hidden targets, employing a DQN-based critic and a coordinator for optimal action selection upon target visibility. Xu et al. \cite{xu2021efficient} adopt a relabeling strategy, recasting mistakenly grasped objects as new targets to optimize training data use, alongside proposing a tri-phase training regime to boost the learning efficiency of pushing and grasping networks. Kiatos et al. \cite{kiatos2019robust} employ a target-centered mesh grid for target state representation, facilitating a push policy aimed at isolating the target by distancing it from nearby objects. Huang et al. \cite{huang2022interleaving} leverage Monte Carlo Tree Search, augmented by a physics engine, to foresee push action outcomes and deduce the most efficient path to the grasp, while incorporating a Push Prediction Network (PPN) to expedite the search. Yu et al. \cite{yu2023iosg} apply a few-shot learning technique to ascertain the target object's probable location, enhancing the identification accuracy amidst similar textures.}

\zlw{Despite these advancements, existing approaches face challenges in achieving motion efficiency within heavily cluttered environments. Consequently, enhancing the capability of household robots to execute goal-oriented grasping in these complex scenarios is crucial.}
\subsection{Amodal Segmentation}
\zlw{Amodal segmentation, a task adept at simultaneously revealing both visible and hidden sections of objects, has been rigorously examined in works such as \cite{li2016amodal,qin2020s4g,back2022unseen}. This capacity to map out occluded areas proves crucial for certain robotic grasping endeavors. Research by Wada et al. \cite{wada2019joint, wada2018instance} utilizes insights from amodal segmentation to decode the layered arrangement of objects, guiding the prioritization of grasps. Further, studies by \cite{yang2019embodied} and \cite{price2019inferring} harness the comprehensive visibility provided by amodal segmentation for engaging in active perception and meticulous object-searching tasks. Nevertheless, the potential of amodal segmentation in the realm of goal-oriented grasping remains largely untapped. Exploring its capabilities within this context holds promise for significant advancements.}
% ...

\section{PROPOSED APPROACH}
\label{sec:method}

In this section, we present our Occluded Push-Grasp Policy (OPG-Policy) approach, which includes an amodal segmentation module to obtain comprehensive information about the target, and a dedicated push-grasp policy learning framework incorporating amodal masks. We detail the procedures of training the amodal segmentation module and the push-grasp policy with amodal masks, including model architectures, reward design, and training techniques.
\subsection{System Overview}
Fig. \ref{fig:pipeline} illustrates the overall pipeline of our method. A fixed-mount RGB-D camera is set to capture the RGB and depth images of the workspace, which are then put into the pre-trained amodal segmentation module to segment the target amodal mask. Afterward, the RGB, depth and target amodal mask images are projected orthographically in the gravity direction to build the color heightmap $I$, depth heightmap $D$, and target amodal mask heightmap $A$, which represent the state $(I, D, A)$. To encode the rotation information into these heightmaps, we rotate them at various angles w.r.t. the z-axis, as detailed in Sec. \ref{subsec:deepQ}. These rotated heightmaps are fed to an encoder to extract the features and then forwarded to the push and grasp Q net to construct the push and grasp Q maps. Optimal pushing and grasping actions are determined by selecting the indexes with the highest Q value in their respective Q maps. Domain knowledge, derived from the task amodal mask, and the best action Q values are input to the coordinator to determine the optimal action type — pushing or grasping. Finally, we construct the push or grasp primitive action and forward it to the robot.

\subsection{Amodal Segmentation}
Amodal segmentation generates amodal masks, which reveal both the visible and occluded parts of target objects. We employ UOAIS \cite{back2022unseen}, a hierarchical feature fusion method, to provide amodal masks for our framework. To collect the training data for this amodal segmentation module, we leverage a general segmentation method (\eg, Segment Anything Model (SAM) \cite{kirillov2023segment}) to segment the visible part of objects in simulation. Additionally, to obtain an amodal mask annotation, we establish a comparison area that matches the size of the workspace. Each time, we place a single object within the comparison area, aligning its pose with the one in the workspace for comparison. Subsequently, SAM \cite{kirillov2023segment} is employed to segment the images of both the workspace and comparison area, producing visible and full masks for the occluded object. We have developed filtering and matching algorithms to synchronize these masks to compute the amodal mask annotations, which serves as training data for amodal segmentation module in our framework.

\subsection{Deep Q-Network}
\label{subsec:deepQ}
This goal-oriented grasping problem can be modeled as a discrete Markov Decision Process (MDP). The state is defined as a tuple $(I, D, A)$, where $I$ represents the color heightmap, $D$ represents the depth heightmap, and $A$ represents the target amodal mask heightmap. All these heightmaps will be rotated by 16 angles (22.5° step size) to incorporate rotational information. Afterward, they are passed to feature extraction module, a feature encoder consists of two convolutional layers followed by a densenet121 pre-trained on ImageNet \cite{deng2009imagenet}. These features are then fed into the grasp Q net and push Q net to generate the pixel-wise Q maps. The push Q net and the grasp Q net share the same structure, each containing three convolutional layers followed by bilinear upsampling. Every pixel value in the Q maps represents the expected future return once the corresponding action is executed in the corresponding position and orientation. This indicates the best pushing and grasping actions are the actions with the highest Q value in Q maps.

\subsection{Coordinator}
The coordinator determines the optimal action type based on the current objects distribution and target state, requiring domain knowledge to make informed decision. It is designed as a three-layer MLP binary classifier for two action types(\ie, pushing and grasping), and it can be formulated as:
\begin{equation}
y=f_{mlp}(q_p,\ q_g,\ o,\ a_b,\ a_n,\ f_c)
\end{equation}
where $q_p$ and $q_g$ are the best push Q value and the best grasp Q value respectively, which represent the action future expected return. $o$ is the target object occluded rate, representing the ratio between its occluded portion and its complete mask. $a_b$ and $a_n$ are indicators of the surroundings of the target object, which can be formulated as:
\begin{equation}
a_b=o_b/t_b, \ a_n=o_b / t_m
\end{equation}
where $o_b, t_b, t_m$ are computed based on the target border $m_b$, a strip that extends 10 pixels outward from the boundary of the target full mask (convert from target amodal mask). $o_b$ denotes the target occluded border value, which corresponds to the occluded portion of $m_b$. $t_b = \sum m_b$, which represents the sum of the target border value. $t_m$ denotes the sum of the target full mask. $a_b$ is the ratio of the occluded target border to full target border, reflecting the target border occlusion situation. $a_n$ is the ratio of the occluded target border to the full target mask, reflecting the proportion of the occluded target border to object size. To some extent, they indicate the occlusion and crowding around the target. $f_c$ denotes the grasp fail count, which reflects the fail history of grasping. These information is fed to the coordinator to determine the optimal action type (\ie, pushing or grasping).

The supervision information for the coordinator is obtained by observing the outcome caused by the selected action. If the coordinator provides a grasping action and it succeeds, a reward of 1 is obtained. Otherwise, the reward is 0. We exclusively utilize information from grasping actions to train our coordinator, as the outcome of a grasping action is easily observable and assessable.

\subsection{Reward}
\subsubsection{Grasping}
For grasping action, efficiently and accurately grasping out the target object is required. To guarantee that grasping accurately locates the position of the target object, we assign a reward of 0.25 when the grasping position falls within the range of the target's amodal mask. Once the target object is successfully grasped out, a reward of 1 is obtained. Therefore, the reward function is formulated as:
\begin{equation}
R=
\left\{
    \begin{array}{ll}
        0.25, & \text{grasp position is within target amodal mask} \\
        1,    & \text{target is grasped} \\
        0,    & \text{otherwise}
    \end{array}
\right.
\end{equation}
\subsubsection{Pushing}

For pushing action, the objective is to clear out the obstacles around the target object to facilitate easier grasp, especially when the target object is heavily occluded. We assign a reward of 0.5 when a pushing action reduces the occluded rate of the target object by 0.1. To increase the success rate of subsequent grasping attempts, a reward of 1 is obtained when a pushing action raises the next best grasp Q value beyond a dynamically adaptive threshold $T_g$, formulated as:
\begin{equation}
T_g = \beta * T_{g} + (1 - \beta) * Q_g
\end{equation}
where $Q_g$ represents the next best grasp Q value after a pushing action is executed. $\beta$ denotes the decay factor, which is set to 0.95. $T_g$ is updated using exponentially weighted updates after each pushing action, as the grasp Q net changes during the training process. Therefore, a dynamically adaptive threshold is established to accommodate these changes. The pushing reward function is formulated as: 
\begin{equation}
R=
\left\{
    \begin{array}{ll}
        0.5,  & \text{target occluded rate decrease over 0.1} \\
        1,    & \text{next best grasp Q value exceed} \ T_g \\
        0,    & \text{otherwise}
    \end{array}
\right.
\end{equation}

\subsection{Training Details}
\zlw{To progressively acclimate the agent to complex scenarios, we systematically escalate the scene's complexity. Initially, for the first 1000 iterations, we introduce a scene with $n=7$ target objects and $m=3$ obstructive objects. Between 1000 and 3000 iterations, the count of obstructions increases from $m=3$ to $m=8$. Subsequently, from 3000 to 5000 iterations, we set the obstacle count to $m=13$. Beyond 5000 iterations, the obstacle count stabilizes at $m=18$,  significantly complicating the agent's ability to grasp the target object successfully.}

\zlw{Additionally, we vary the occlusion states of objects across iterations for diversified learning experiences. Up to 3000 iterations, target objects are selected at random. Post-3000 iterations, preference is given to the most occluded object as the target, aiming to enhance the agent's proficiency in more demanding scenarios. The agent is tasked with retrieving all potential target objects from the scene before it is reset with a new set of objects, preventing the agent from becoming mired in complex scenarios.}

\zlw{During the initial 1000 iterations, we employ an $\epsilon$-greedy policy for action selection instead of the coordinator, acknowledging the agent's nascent grasping skills at this stage. Employing a coordinator at this stage could lead to wrong action choices. However, as the agent's ability to grasp improves post-1000 iterations, we switch to using the coordinator for action type selection, aligning the strategy with the agent's enhanced capabilities.}
\zlw{The training of the DQN model involves minimizing the temporal difference error $\delta_t$, which is expressed as follows:}
\begin{equation}
\delta_t = Q(s_t,a_t;\theta_t) - [R_t(s_t,a_t,s_{t+1})+\gamma \cdot \max_q(s_{t+1},a_{t+1};\theta_t)]
\end{equation}
where $s_t$ demotes the state at time $t$, $a_t$ represents the action taken at time $t$, and $\theta_t$ are the network parameters at time t. $\gamma$ is the future discount factor, a constant of 0.5. $s_{t+1}$ and $a_{t+1}$ is the state and action at next time $t+1$. Upon calculating out the $\delta_t$, we employ the Huber loss function to define our loss, which is formulated as:
\begin{equation}
L = 
\begin{cases} 
\frac{1}{2}\delta_t^2, & \text{if } |\delta_t| \leq 1 \\
|\delta_t| - \frac{1}{2}, & \text{otherwise}
\end{cases}
\end{equation}
We use binary cross-entropy loss to train the coordinator, which is formulated as:
\begin{equation}
L = -(\overline y \log y + (1 - \overline y) \log (1 - y))
\end{equation}
where $y$ is the output of the coordinator, and $\overline y$ is the ground-truth(\ie, successful grasp is represented as 1, while a failure is represented as 0).

\begin{figure}[t]
\begin{center}
\includegraphics[width=\linewidth]{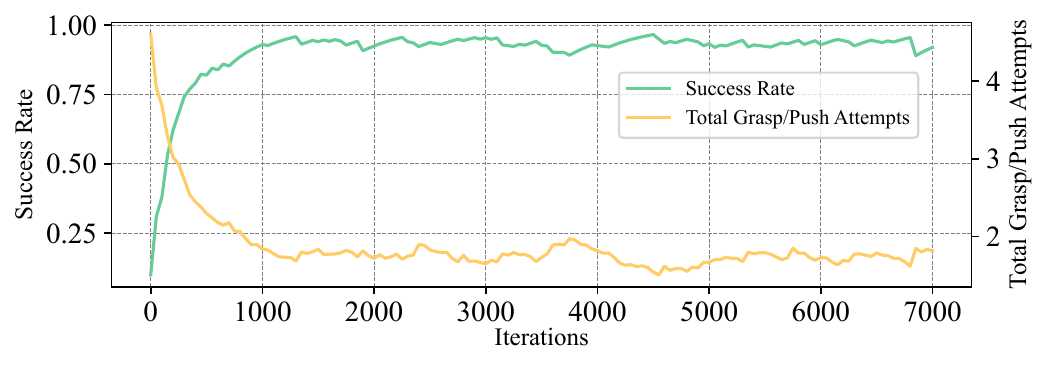}
\end{center}
\vspace{-15pt}
\caption{\textbf{Training performance.} The green and yellow lines respectively indicate the changes in the success rate and total grasp/push attempts of our model during the training process, based on measurements of 15 objects.}
\vspace{-10pt}
\label{train_motion_success}
\end{figure}

The training process is depicted in Fig. \ref{train_motion_success}. It is evident that as the number of training iterations increases, there is a notable improvement in the success rate for completing 15 object retrieval tasks, accompanied by a reduction in the total number of grasp/push attempts. This trend underscores the effectiveness of our training methodology.

\begin{table*}[!t]
\centering
\caption{Simulation Results on Random Test}
\label{Tab: random test}
\begin{tabular}{c|c|c|c|c|c|c|c|c}
\hline \multirow{2}{*}{ Methods } & \multicolumn{4}{c|}{Success Rate  (\%) ↑ } & \multicolumn{4}{c}{ Total Grasp/Push Attempts ↓ } \\
\cline { 2 - 9 } & 15 objects & 30 objects & 30 objects (\textit{hard}) & Average & 15 objects & 30 objects & 30 objects (\textit{hard}) & Average \\
\hline
GTI \cite{yang2020deep} & 88.00 & 73.00 & 33.00 & 64.66 & 2.02 & 3.03 & 4.26 & 3.10\\
GTI-amodal* & 71.00 & 53.00 & 27.00 & 50.33 & 2.4 & 3.13 & 4.51 & 3.34\\
GE-GRASP \cite{liu2022ge} & \textbf{97.00} & 81.00 & 57.00 & 78.33 & \textbf{1.66} &  2.59  & 3.98 & 2.74\\
OPG-Policy-no-coordinator* & 96.00 & 77.00 & 58.00 & 77.00 & 1.85 & 2.47 & 3.71 & 2.67\\
OPG-Policy &  96.00  &  \textbf{84.00}  &  \textbf{68.00} & \textbf{82.66}  & 1.79 & \textbf{2.38} & \textbf{3.69} & \textbf{2.61} \\
\hline
\end{tabular}
\vspace{1pt}
\begin{tablenotes}
\item * \textbf{GTI-amodal} is an adaptation of the GTI method, where the visible segmentation module is replaced with our amodal segmentation module. \textbf{OPG-Policy-no-coordinator} is a variant of our method that omits coordination, selecting actions based solely on the highest Q-value.
\vspace{-10pt}
\end{tablenotes}
\end{table*}

\section{EXPERIMENTS}

\zlw{In this section, we comprehensively evaluate our method through simulation experiments and real-world applications, focusing on three main objectives: validating the efficiency of object retrieval from a cluttered workspace, demonstrating superior performance compared to existing baseline methods, especially in scenarios with heavy occlusion of the target object, and assessing the system's ability to be directly applicable in real-world settings from simulation training without the need for fine-tuning.}

\begin{figure}[t]
\begin{center}
\includegraphics[width=\linewidth]{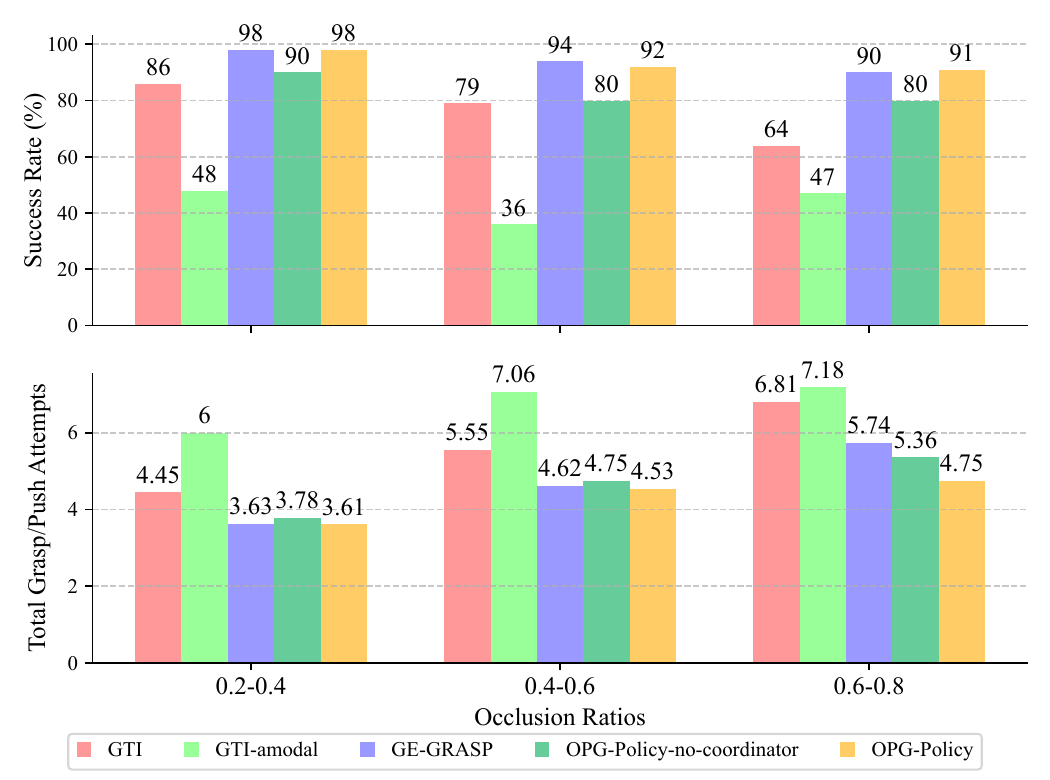}
\end{center}
\vspace{-15pt}
\caption{\textbf{Performance on different occlusion ratios test.} The task success rate (upper) and total grasp/push attempts (lower) of five approaches on different occlusion ratios test. Our method outperforms other methods by at least 1\% in success rate and by at least 0.61 in total grasp/push attempts in the most complex case (0.6-0.8 occlusion).}
\vspace{-15pt}
\label{fig:occlusion test}
\end{figure}

\subsection{Simulation Experiments}
\zlw{In this section, we undertake a performance comparison of our method against benchmarks within a simulated environment. To ensure a fair assessment, we adhere to the simulation parameters established by \cite{yang2020deep}, utilizing a simulated UR5 robotic arm equipped with an RG2 gripper as modeled in V-REP \cite{rohmer2013v}. Our analysis spans three scenario types—random, occluded, and challenging—to rigorously evaluate and offer a comprehensive comparison of the different methodologies involved.}



\begin{figure*}[t]
\begin{center}
\includegraphics[width=\linewidth]{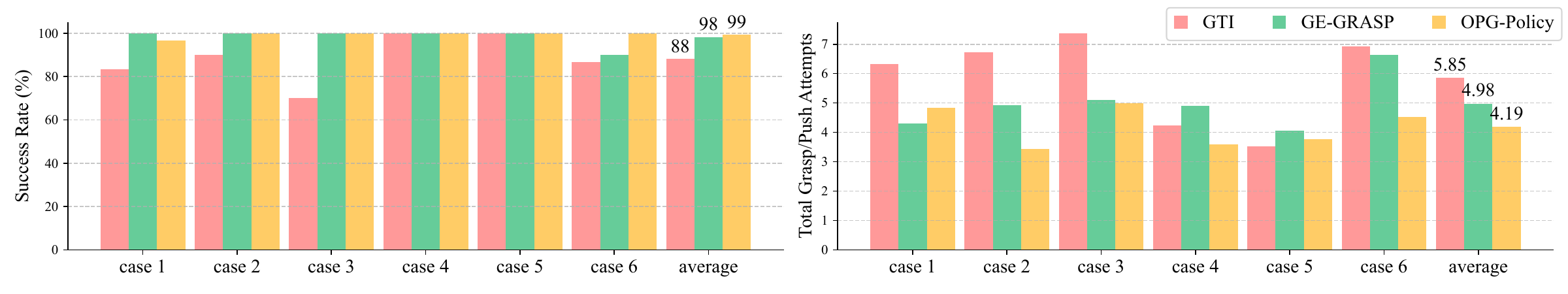}
\end{center}
\vspace{-15pt}
\caption{\textbf{Performance on challenging occlusion cases}, specifically success rate and total grasp/push attempts for each method. Our method leads the second-place method \textbf{GE-GRASP \cite{liu2022ge}} by 1\% in terms of success rate and by 0.79 in terms of total grasp/push attempts on average.}
\vspace{-15pt}
\label{fig:chanllenge test}
\end{figure*}

\subsubsection{\textbf{Random Case}}

In the tests, we set up three scenarios containing 15 or 30 objects where the target is randomly selected, and 30 objects where the heaviest occluded object is selected as the target (named as 30 objects (hard)). The complexity of a scene increases with the number of objects increasing from 15 to 30. In different scenarios, the specified number of objects of random colors and shapes are dropped into the workspace from randomly selected positions and orientations. Once an object is chosen as the target, the others become obstacles. If the target is retrieved, 5 grasp and push attempts are conducted, or 5 consecutive failures happen, a trial is considered to be finished and the scene will be reset. To ensure a fair comparison, the same random seed is used for each method. 100 independent trials are performed on each scenario for each method, and the results are shown in Table I. In the scenario with 15 objects (the target randomly selected), \textbf{GE-GRASP} performs the best by a slight advantage. However, when the number of the objects increases (30 objects) or/and the heaviest occluded object is selected as the target (30 objects (hard)), our system remarkably outperforms the other strategies both in success rate and motion efficiency. In addition, our approach achieves the highest average success rate of 82.66\% and the lowest average total grasp/push attempts of 2.61. It is noted that despite \textbf{GTI-amodal} employing amodal masks, its performance remains worse than \textbf{GTI}. It indicates that simply integrating the amodal masks into the \textbf{GTI} may even degrade the performance. Meanwhile, as shown in Table I, our system outperforms \textbf{OPG-Policy-no-coordinator} with respect to the average success rate as well as the total grasp/push attempts. It highlights the significance of the coordinator in the proposed approach to generate the robot’s optimal motion.

\begin{figure}
\centering
\includegraphics[width=0.9\linewidth]{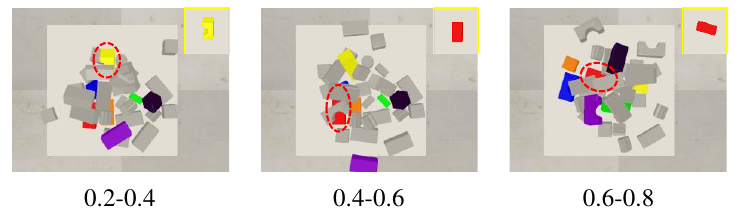}
\caption{\textbf{Examples of different occlusion ratio cases in simulation.} These three cases represent occlusion ratios of 0.2-0.4 (left, yellow block), 0.4-0.6 (center, red block), 0.6-0.8 (right, red block), respectively. They are randomly selected from blocks that meet the requirements of the occlusion ratio.}
\label{fig:occlusion}
\vspace{-8pt}
\end{figure}

\subsubsection{\textbf{Occluded Case}}
\zlw{To investigate the retrieval efficiency of occluded target objects, we design occlusion experiments with objects featuring varied occlusion ratios. Figure~\ref{fig:occlusion} segments these ratios into three groups: 0.2-0.4 (mild occlusion), 0.4-0.6 (moderate occlusion), and 0.6-0.8 (severe occlusion), based on the extent of the target object's mask that is occluded relative to its full mask. This setup aligns with our random case, maintaining a constant number of 100 trials, 30 objects, and the maximum total push/grasp attempts are set as 10 across all scenarios. As illustrated in Figure~\ref{fig:occlusion test}, our system demonstrates a competitive success rate against \textbf{GE-GRASP} in scenarios of severe occlusion. Nevertheless, it exhibits superior motion efficiency by reducing the average number of grasp/push attempts by nearly one. This improvement is attributed to our method's advanced understanding of the target object's state via amodal segmentation, enhancing our Q-networks' ability to devise more effective action strategies compared to the more limited perception model of \textbf{GE-GRASP}. It is also observed that \textbf{OPG-Policy-no-coordinator} secures a secondary position in terms of efficiency, benefiting from amodal segmentation. However, its success rate is compromised due to the absence of a coordination mechanism, leading to repetitive and inefficient grasping actions.}

\begin{figure}
    \centering
    \includegraphics[width=0.9\linewidth]{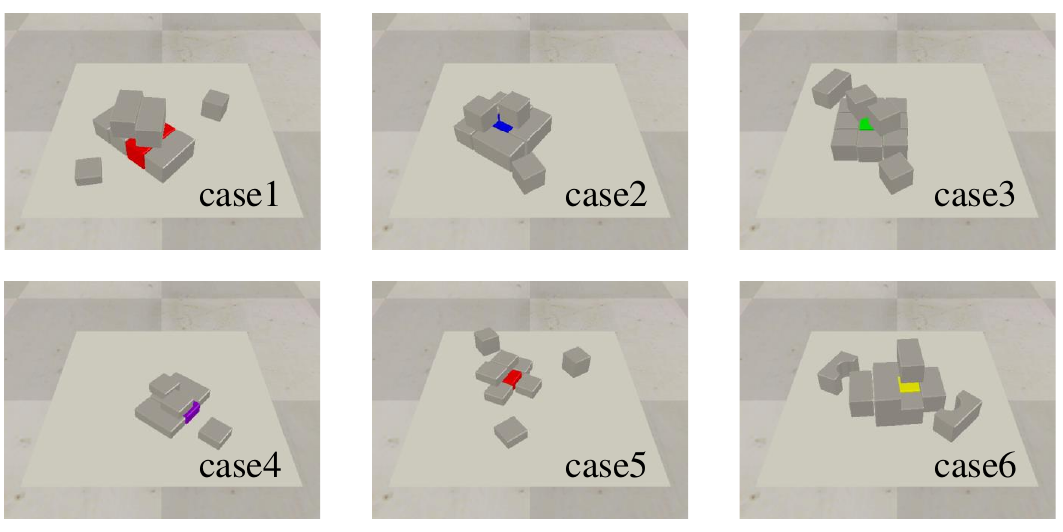}
    \caption{\textbf{Challenging occlusion cases in simulation.} These six cases are constructed to evaluate the ability of different methods to grasp occluded targets in structured environment. The colored objects represent the targets.}
    \vspace{-5pt}
    \label{fig:all}
\end{figure}

\subsubsection{\textbf{Challenging Case}}



\zlw{Evaluating our coordination capabilities involved designing six scenarios (illustrated in Figure \ref{fig:all}) inspired by \cite{yang2020deep}.  We increased the complexity by introducing extra obstacles above the target object, thereby inducing occlusion. These scenarios are carefully crafted to simulate environments where objects are not only confined by their surroundings but are also obscured by occlusions. Each scenario undergoes 30 repeated trials and \dingh{the maximum total grasp and push attempts are set as 10} per trial. The results, as shown in Figure \ref{fig:chanllenge test}, indicate the superior performance of our method in most scenarios. On average, across all scenarios, our approach achieves a 99\% success rate and a movement count of 4.19, significantly outperforming \textbf{GTI}, which achieves an 88\% success rate and 5.85 movements, and \textbf{GE-GRASP}, with a 98\% success rate and 4.98 movements.}
\vspace{-5pt}
\subsection{Real World Experiments}

\zlw{Our real-world experimentation employs a UR10 
robotic arm equipped with a ROBOTIQ 
gripper, facilitating nuanced tabletop movements. An Intel RealSense D435i camera, affixed to the gripper, captures RGB-D images at a resolution of 640x480, ensuring a consistent viewpoint and orientation. Remarkably, the model deployed in these real-world tests is trained exclusively in a simulated environment, without necessitating any real-world fine-tuning. GTI \cite{yang2020deep} and GE-GRASP \cite{liu2022ge} are selected as benchmarks for comparative analysis. To maintain a level playing field, especially considering the inadequate performance of the built-in perception modules of GTI and GE-GRASP in real-world settings, we integrate UOAIS \cite{back2022unseen} to supply enhanced quality visible masks for these baselines. In our generalization assessments, UOAIS, trained on the YCB \cite{calli2015benchmarking} object dataset within a simulated framework, is utilized to generate both amodal and visible masks. Consequently, our selection of target objects is exclusively limited to those from the YCB dataset.}


\zlw{The evaluation encompasses two primary components: challenging tests and generalization tests. In the challenging component, four distinct scenarios are crafted, each placing the target object amidst and partially concealed by surrounding items. We undertake 10 trials for each scenario, \dingh{and maximum the total grasp and push attempts are set as 10}. The generalization component, on the other hand, is designed to test the model's ability to adapt to new environments. It does so by populating the workspace with twelve household objects that are unfamiliar to the models from their training. Each trial within this component selects one target object from the YCB \cite{calli2015benchmarking} dataset present in the scene. Upon completion of a trial, the total number of objects is reset to twelve, and their placement is randomized for the subsequent trial. This procedure is repeated across 20 generalization trials for each method, with \dingh{the maximum total grasp and push attempts set as 10} for each trial.}

\zlw{The result summarized in Table \ref{Tab: real test} indicates that \dingh{our system} excels in the challenging tests, leading the next best method by 2.5\% in success rate and by 0.62 in total grasp / push attempts. Furthermore, in the generalization test, \dingh{our system} shows its strength in adaptability, outperforming the second-best method by a 5\% margin in success rate and by 0.65 in total grasp/push attempts. These results highlight \dingh{our system} exceptional ability to generalize across different settings.}


\begin{figure}
\centering
\includegraphics[width=\linewidth]{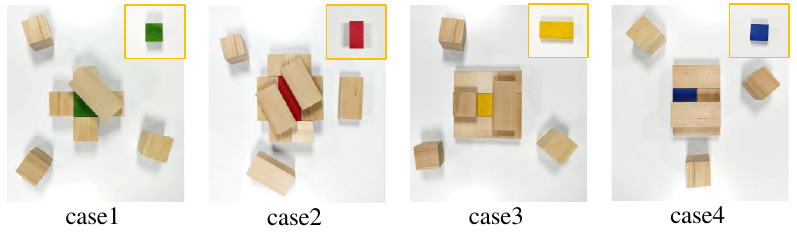}
\caption{\textbf{Challenging test on real robot.} In each of these four test cases, there is a target object that is occluded and surrounded by other objects, posing a challenging situation.}
\vspace{-10pt}
\label{fig:c-test}
\end{figure}

\begin{figure}
\centering
\includegraphics[width=0.75\linewidth]{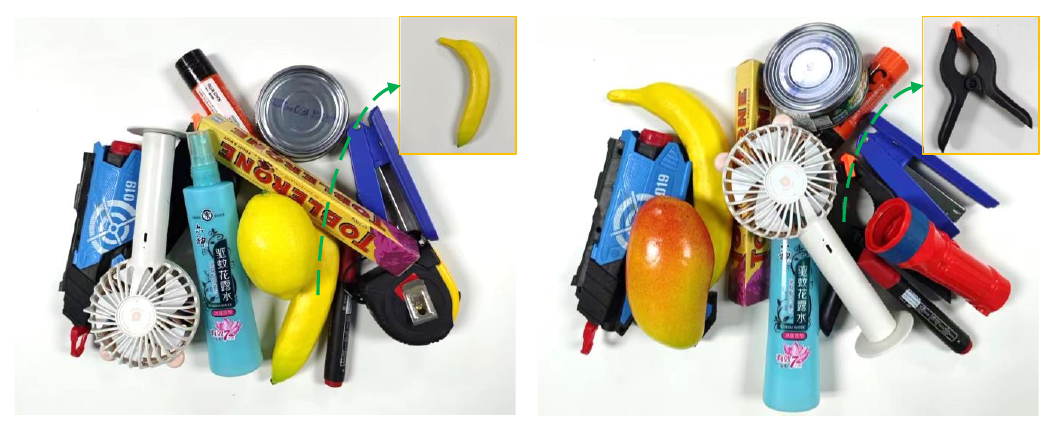}
\caption{\textbf{Generalization test on real robot.} Two examples consist of everyday objects, which used to evaluate the generalization of our method to the daily objects.}
\vspace{-15pt}
\label{fig:g-test}
\end{figure}

\begin{table}[t]
\centering
\caption{Real-world Test Results}
\label{Tab: real test}
\begin{tabular}{c|c|c|c|c}
\hline \multirow{2}{*}{Methods} & \multicolumn{2}{c|}{Success Rate (\%) ↑ } & \multicolumn{2}{c}{Total Grasp/Push Attempts ↓ } \\
\cline{2-5} & C-Test & G-Test & C-Test & G-Test \\
\hline
GTI & 65.00 & 80.00 & 8.42 & 7.75 \\
GE-Grasp & 82.50 & 85.00 & 7.62 & 7.3 \\
OPG-Policy & \textbf{85.00} & \textbf{90.00} & \textbf{7.0} & \textbf{6.65} \\
\hline
\end{tabular}
\begin{tablenotes}
\item * C-Test and G-Test stand for Challenging Test and Generalization Test respectively.
\vspace{-10pt}
\end{tablenotes}
\end{table}

\section{CONCLUSIONS}

In this work, we propose Occluded Push-Grasp Policy (OPG-Policy), the first approach to introduce amodal masks into push-grasp policy learning framework. Specifically, we begin by training an amodal segmentation module and then design a specific reward and training stages, including coordinator, to incorporate the additional knowledge from amodal masks into training the push Q net and grasp Q net, as well as determining the optimal action. This approach leverages the prior knowledge of target objects to facilitate the policy in generating optimal actions, effectively improving motion efficiency. Our results, including real-world experiments, outperform other competitive baselines in more complex scenarios. Additionally, our further analysis indicates that the amodal segmentation module and coordinator play significant roles in our method.

\label{sec:subjective}





{
\bibliographystyle{IEEEtran}
\bibliography{IEEEabrv,reference}
}






\end{document}